\documentclass[10pt,twocolumn,letterpaper]{article}

\usepackage{iccv}
\usepackage{times}
\usepackage{epsfig}
\usepackage{graphicx}
\usepackage{amsmath}
\usepackage{amssymb}
\usepackage{subfigure}

\usepackage[breaklinks=true,bookmarks=false]{hyperref}

\iccvfinalcopy 

\def\iccvPaperID{****} 

\begin{document}

\title{
Relaxed Spatio-Temporal Deep Feature Aggregation for Real-Fake Expression Prediction
}

\author{
Savas Ozkan$^{1,2}$, Gozde Bozdagi Akar$^{2}$\\
$^1$ TUBITAK UZAY, Image Processing Group, Ankara, Turkey\\
$^2$ Middle East Technical University, Multimedia Lab., Ankara, Turkey\\
{\tt\small savas.ozkan@tubitak.gov.tr, bozdagi@metu.edu.tr}\\ \\
\small \url{https://github.com/savasozkan/real-fake-emotions}
}

\maketitle

\begin{abstract}
Frame-level visual features are generally aggregated in time with the techniques such as LSTM, Fisher Vectors, NetVLAD etc. to produce a robust video-level representation. We here introduce a learnable aggregation technique whose primary objective is to retain short-time temporal structure between frame-level features and their spatial interdependencies in the representation. Also, it can be easily adapted to the cases where there have very scarce training samples. We evaluate the method on a real-fake expression prediction dataset to demonstrate its superiority. Our method obtains $65\%$ score on the test dataset in the official MAP evaluation and there is only one misclassified decision with the best reported result in the Chalearn Challenge (i.e. $66.7\%$) . Lastly, we believe that this method can be extended to different problems such as action/event recognition in future.


\end{abstract}

\section{Introduction}
Nowadays, video is an essential multimedia resource that has critical applications on several domains including surveillance, entertainment, social networking and event prediction. Even though their high popularity and information capacity, it is really hard to understand and monitor the visual content of a video by computers. In recent years, the success on deep learning (especially on end-to-end feature learning) accelerates the progress and the research efforts on video understanding. 

In literature, common techniques are generally based on the aggregation of frame-level deep features computed on a pretrained convolutional neural network (CNN) model~\cite{xu2015discriminative, girdhar2017actionvlad, donahue2015long, gao2016compact, arandjelovic2016netvlad}. However, even if the techniques modeling only the spatial  interdependencies of deep features attain superior performances for large-scale data, they are not able to capture the temporal structure between these features~\cite{xu2015discriminative, arandjelovic2016netvlad}. Similarly, pooling techniques based on temporal models~\cite{donahue2015long, hochreiter1997long} are inclined to overfit to the training data and irrelevant results can be obtained for unseen examples. Recently,~\cite{miech2017learnable} shows that random frame sampling yields similar results for the temporal models compared to dense frame sequences. This is a critical observation for which temporal models have some difficulties to learn/generalize complex spatio-temporal structures between actions from coarse frame-level features.

\begin{figure}[t]
\centering
\subfigure{
\label{fig:a}
\includegraphics[scale=0.058]{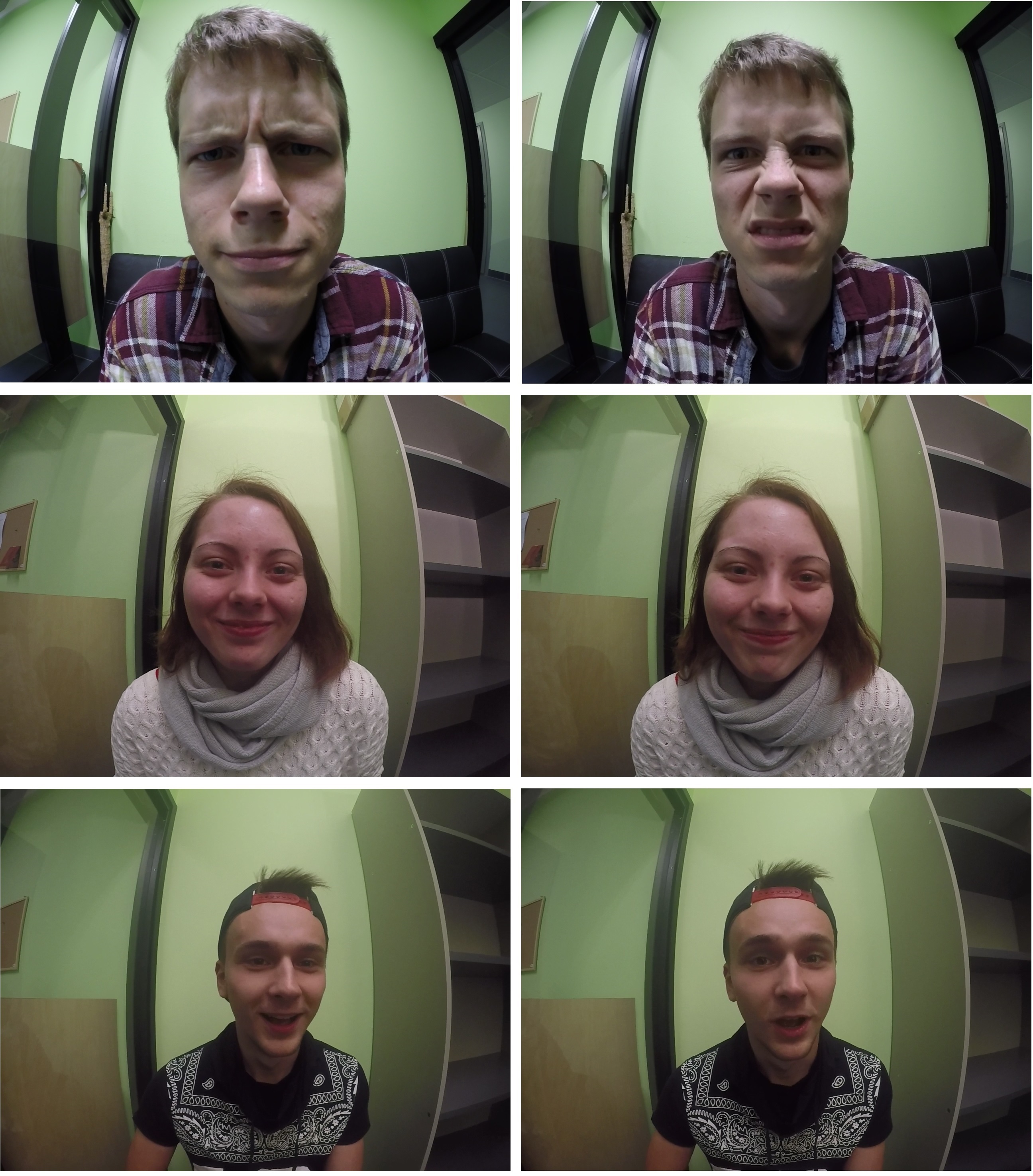}}
\subfigure{
\label{fig:b}
\includegraphics[scale=0.058]{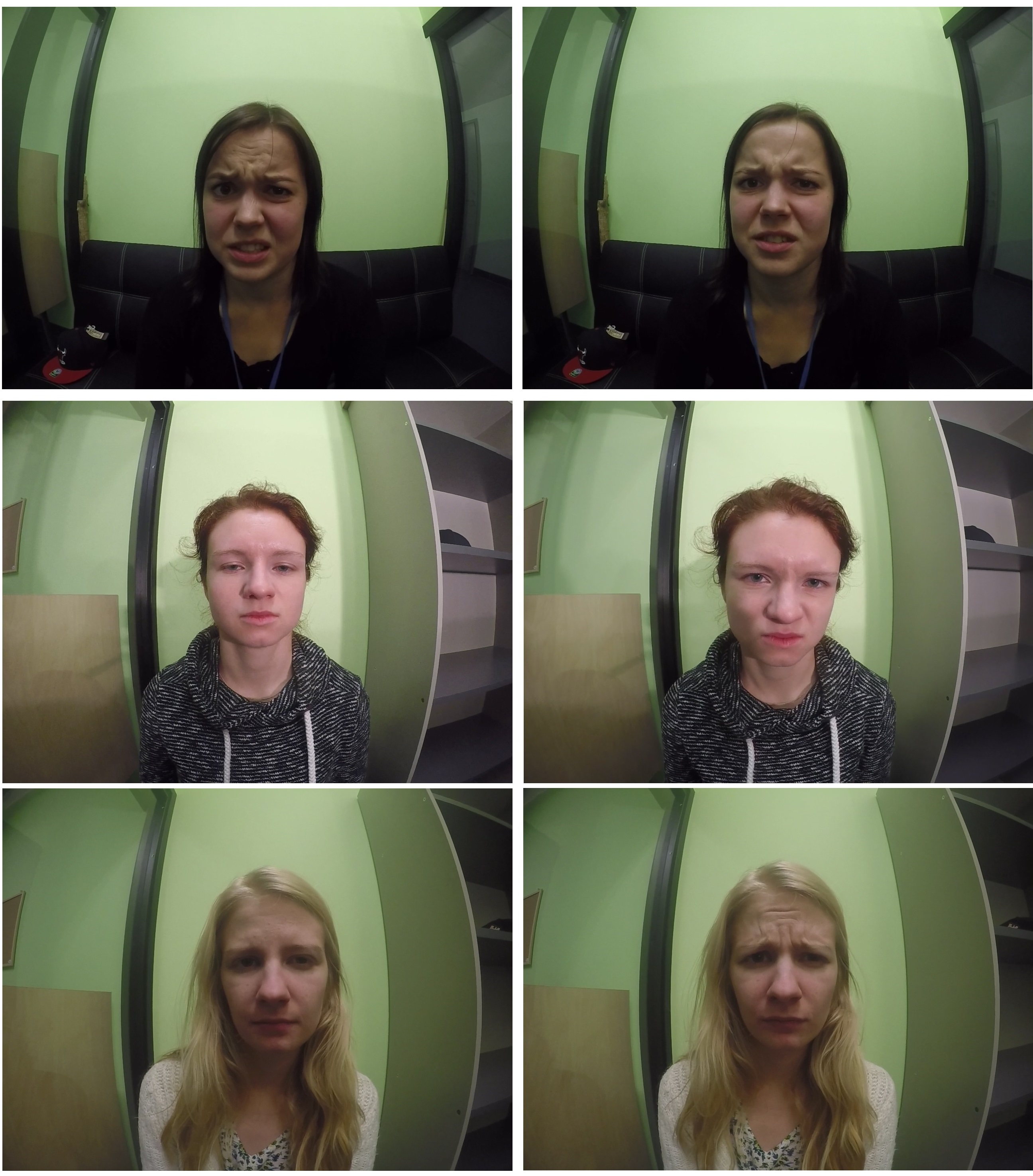}}
\caption{Some visual examples for genuine (first image) and deceptive (second image) expressions. }
\label{fig:data}
\end{figure}

\begin{figure*}[t]
\centering
\includegraphics[scale=0.3]{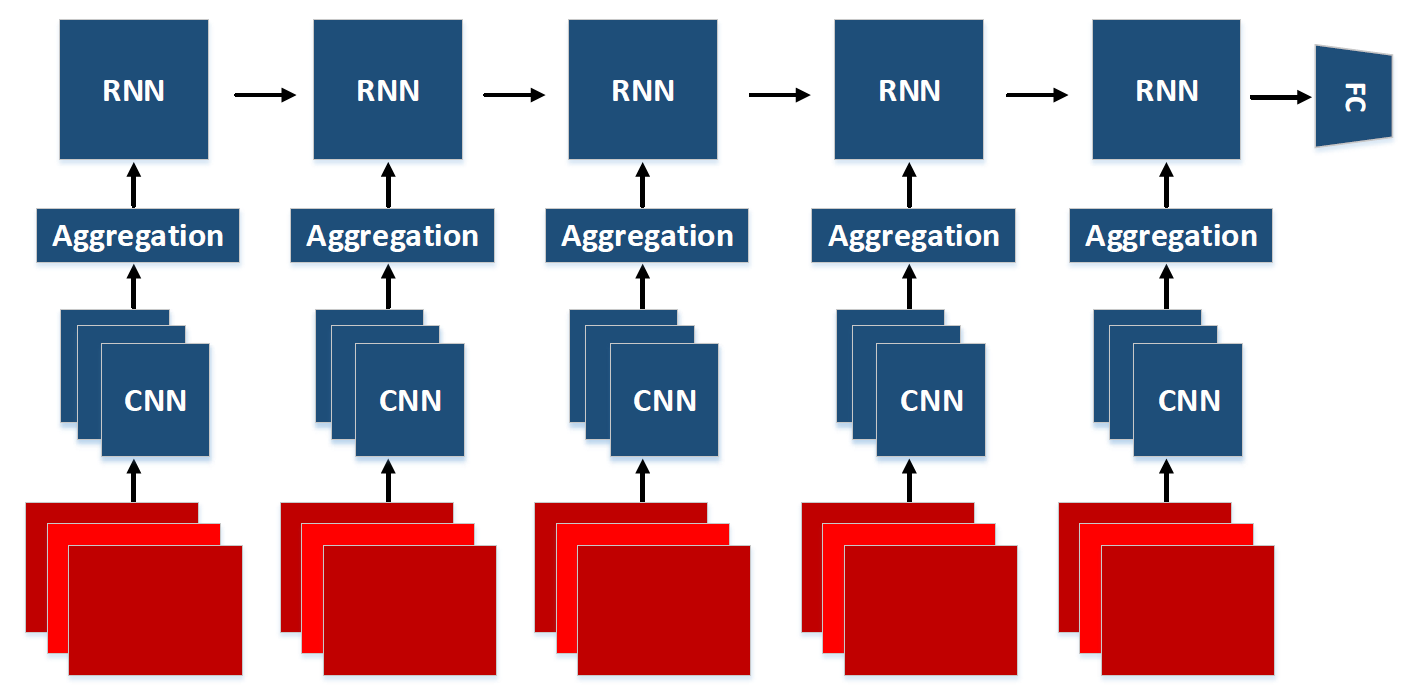}
\caption{Flow of relaxed spatio-temporal feature aggregation method. Initially, frame-level deep features are extracted and aggregated from multiple-frames (red). Later, temporal structure is captured by RNN. Particularly, this architecture is more suitable for the cases where there have very scarse training samples.}
\label{fig:flow}
\end{figure*}

For facial emotions, real-fake expression prediction on a video is slightly different from frame-level emotion classification: 1) Temporal interrelations between face parts should be considered, since temporal consistency is highly critical to categorize human expressions~\cite{trutoiu2014spatial}. 2) Emotional changes in eyes and mount movements can be distinct to separate real expressions from fake ones~\cite{iwasaki2016hiding}. 

In this paper, we introduce a learnable aggregation method that is able to capture the short-time spatio-temporal structure between deep visual features simultaneously. This is particularly important for real-fake expression prediction, since learning micro-emotional interdependencies from face parts as well as their sequential structure can be essential to distinguish the real-fake expressions as we mentioned. To advance the performance, we use high-level convolution features (conv5) of CNN that are finetuned for emotion detection~\cite{ng2015deep}. 

Another contribution is that this model can be easily adapted to the cases where there have very scarce training samples. This can be achieved by only replacing the trainable pooling parts in the architecture with compact ones or vice versa.

\section{Related Work}
We review video representation and facial expression prediction under two different subsections since there are interrelated for real-fake expression prediction as follows.

\vspace{0.2cm}
\noindent
\textbf{Video Representation}: Initial attempts to encode visual content of a video are broadly formulated with hand-crafted features. These features are extracted from either local patches~\cite{scovanner20073, snoek2009mediamill} and/or local trajectories~\cite{wang2011action} by exploiting regional gradient features. Later, they are aggregated over frames with various pooling techniques to obtain a single compact representation~\cite{perronnin2010improving, jegou2010aggregating}. Recently, feature extraction and aggregation steps are adapted to learnable modules. Frame-level~\cite{xu2015discriminative, donahue2015long}, spatio-temporal-level~\cite{varol2017long, ji20133d} and motion-level~\cite{simonyan2014two} convolutions are extensively studied to extract robust features from videos. Moreover, trainable pooling techniques are introduced ~\cite{girdhar2017actionvlad, arandjelovic2016netvlad} to boost discriminative representation learning rather than unsupervised feature-space clustering~\cite{perronnin2010improving, jegou2010aggregating}.   

\vspace{0.2cm}
\noindent
\textbf{Facial Expression Prediction}: Same trend can be observed in facial expression prediction similar to video representation. At first, hand-crafted and geometrical features are used to predict emotions from images/videos~\cite{gunes2007bi, pantic2007human}. Later, deep learning techniques are applied to obtain more robust visual descriptions~\cite{ding2017facenet2expnet, ng2015deep, gurpinar2016multimodal, guccluturk2016deep}. However, the limited number of training samples is a major concern on this domain. Thus, studies are usually concentrated on transfer learning and data augmentation to extract reliable deep features by adapting pretrained models to the problem~\cite{ding2017facenet2expnet, ng2015deep}. Also, video-level representation can be obtained by considering simple feature distribution characteristics in time (i.e. mean and variance)~\cite{gurpinar2016multimodal} as well as by common pooling techniques. 

Even if there are vast numbers of studies on facial emotion detection, the prediction of real-fake expression is limited and most of which are for still images~\cite{littlewort2009automatic}. Also, they are usually based on hand-crafted features. Therefore, reviewing the literature from the perspective of cognitive science might be more informative in order to understand and propose a proper technique for real-fake expression prediction. In particular,~\cite{iwasaki2016hiding, trutoiu2014spatial} show that small emotional changes in eyes and mount movements can be used to interpret an expression as genuine (real) or deceptive (fake) from the sequence of faces. This indicates that both spatial interdependencies of face parts and their sequential structure should be considered in the final model. 

\section{Overall Architecture Used for Real-Fake Expression Prediction}

\vspace{0.2cm}
\noindent
The overall architecture consists of two main modules, namely feature extraction and classification. In the first module, we initially extract robust micro-emotional visual representations for each individual emotion type. For this purpose, we propose a learnable method which exploits a pooling technique~\cite{gao2016compact} and a temporal model~\cite{hochreiter1997long} rather than using raw emotion features~\cite{ding2017facenet2expnet, ng2015deep} or feature pooling techniques~\cite{arandjelovic2016netvlad} only to represent a video. Later, these short-time micro-emotional features are aggregated in time and mapped to a single representation for each video. Finally, video-level representations are trained with a linear kernel Support Vector Machine (SVM)\footnote{LibSVM library. \url{https://www.csie.ntu.edu.tw/~cjlin/libsvm/}} for two class prediction problem (i.e. real or fake).
 
\vspace{0.2cm}
\noindent
\textbf{Feature Extraction}: We initially compute high-level emotional features (conv5) by using the pretrained CNN model~\cite{ng2015deep} on the faces\footnote{Dlib library is used for face detection. \url{http://dlib.net/}} extracted from every video frame. In particular, we use the outputs of ReLU activations of conv5 layers as deep features to impose sparsity on them.

Moreover, we observed that the pretrained CNN model still has an undesired bias for similar faces. For this purpose, we separately normalize the deep features (conv5) for each person by subtracting the average of his/her facial emotions during video rather than overall dataset mean. This step ultimately alleviates the oscillation and centers the data to zero mean for learning phase.

Later, our relaxed spatio-temporal aggregation method (we will explain the details of the method in Section 4) takes these deep features as input. Since the real-fake expression dataset has a limited number of samples, we select dense and short-time temporal intervals from videos to augment data (i.e. the length of each interval is set to $150$ ms) in order to learn micro-emotional representations. After learning optimum parameter sets from the training data, spatio-temporal representations are computed from dense and overlapping intervals, and they are aggregated with~\cite{gao2016compact} to obtain a single video-level representation.

Lastly, we apply power normalization (i.e. $\sigma= 0.5$) to each element of the spatio-temporal representation as described in~\cite{gao2016compact, perronnin2010improving}. This eventually decreases the sparsity of the final representation and meaningful classification results can be achieved with the classifiers based on inner-product~\cite{perronnin2010improving}.

\vspace{0.2cm}
\noindent
\textbf{Classification}: For classification module, Support Vector Machine (SVM) with a linear kernel is utilized. $C$ value is fixed to $1$ for all emotion types which yields the best overall accuracy on the validation dataset. Note that in the final evaluation, each emotion type is considered separately and individual real/fake decisions are computed by each classifier.

\section{Relaxed Spatio-Temporal Deep Feature Aggregation}

In this section, we will explain the relaxed spatio-temporal aggregation method introduced in the paper. Briefly, our method can be summarized as the composition of two feature pooling stages and a temporal model, and Fig.~\ref{fig:flow} illustrates the flow of the method.  

First, local deep features are aggregated from short frame grids. Later, a learnable temporal model is utilized to obtain the temporal structure information. Lastly, the outputs of the method are encoded to achieve a compact video-level representation.

Our method differs from the literature~\cite{varol2017long, donahue2015long, ji20133d} with its complete feature pooling assumption (i.e. efficient and effective). Similarly, in~\cite{varol2017long}, local features are encoded on small spatio-temporal grids and temporal structure is captured by a fully connected layer. But, these models need large-scale data to train a reliable model and to find an optimum solution for the problem. On the other hand, our method can still attain moderate performance by merely encoding deep features extracted from a pretrained model while capturing short-time temporal structure. Moreover, it allows to refine the parameters of the CNN model by back-propagating error through the network for more correlated visual deep features in the problem.

\vspace{0.2cm}
\noindent
\textbf{Capturing Interdependencies between Deep Features from Multiple Frames}: Exploiting local feature dependencies with others is an important visual clue to identify the event, action and object from visual data. For instance, building a model that can recognize bird parts from a scene such as 'beak', 'tail' and 'wing' can yield a high confidence for the bird classification problem or even for fine-grained category estimation~\cite{reed2016learning, gao2016compact}. Therefore, encoding local interdependencies between deep features is essential to obtain a robust content representation. In our model, this step is extended over multiple frames to extract more reliable relational descriptions. 

For this purpose, individual local deep features ($\mathbf{x}^1_t, \mathbf{x}^2_t, ..., \mathbf{x}^M_t, ..., \mathbf{x}^1_{t+T}, \mathbf{x}^2_{t+T},... , \mathbf{x}^M_{t+T}$) are aggregated from a regular frame grid in time (i.e. from multiple-frames). This step estimates a frame grid representation $\mathbf{y}_k$ that captures the spatial interdependencies between features for a short interval. Here, $M$ and $T$ denote the number of features per frame and the temporal frame length of a grid. 

Since multiple frames are exploited to find a relaxed short-time representation, it intuitively leaves an error margin for the temporal structure estimation and leads to a more stable temporal representation. More precisely, instead of using individual frame information in a temporal model as in~\cite{donahue2015long}, our method leverages a pooling scheme before a temporal model to capture the interdependencies of deep features on multiple video frame (Note that~\cite{wang2011action} uses the similar assumption as the sum of trajectory features in time and the authors claim that it improves the accuracy). Thus, this decreases the tendency of temporal models to overfit to the training data. 

In the feature pooling step, one of the aforementioned methods can be used such as NetVlad~\cite{arandjelovic2016netvlad} or Compact Bilinear Pooling (CBP)~\cite{gao2016compact}. In particular, we used CBP in this study due to a small number of training samples.

\vspace{0.2cm}
\noindent
\textbf{Capturing Short-Time Temporal Structure}: In order to model the temporal structure, we encode the grid representations with a learnable temporal model. For this purpose, RNN architecture is utilized. This ultimately captures the relaxed temporal structure over the computed representations among grids. 

As mentioned, this architecture (i.e. fusion of relaxed deep feature pooling on multiple-frames and temporal structure) provides several advantages: 1) compared to frame-level feature pooling techniques, it can preserve the partial temporal structure between deep features in the representation. 2) due to the relaxed multi-frame feature pooling mechanism, the drawback of parameter overfitting can be mitigated for RNN architectures. 3) on the contrary to spatio-temporal convolutional methods, it can yield better accuracy in the cases where insufficient samples exist.  

Formally, the temporal model takes $K$ number of grid representations ($\mathbf{y}_1, \mathbf{y}_2,..., \mathbf{y}_{K}$) as input. Since RNN is used, it sequentially processes current input $\mathbf{y}_k$ and RNN cell state $\mathbf{c}_{k-1}$ as a function of $f$ to predict the output of cell $\mathbf{o}_k $ and the next cell state $\mathbf{c}_k$ as below:
\begin{eqnarray}
\label{eqn:rnn1}
\mathbf{o}_k, \mathbf{c}_k = f(\mathbf{y}_k, \mathbf{c}_{k-1})
\end{eqnarray}

\noindent
Here, the final output of RNN architecture, $\mathbf{o}_K$, is used as the representation of the temporal interval.

 Eventually, this method produces a representation within $K \times T$ frame interval. For a video-level representation, this model should be applied to whole video by uniformly or randomly selecting intervals. Lastly, these spatio-temporal representations should be encoded for the final video-level representation. 

\vspace{0.2cm}
\noindent
\textbf{Estimation of Video-Level Representation}: The spatio-temporal representations can be mapped to a compact video representation in different ways by using averaging/max-pooling and feature pooling schemes. The straightforward solution is to encode them with a compact pooling technique. Similarly, NetVlad and CBP can be used, in general, which are suitable for back-propagation to the deeper layers. Again, CBP is employed to reduce the number of trainable parameters in our model for real-fake expression prediction.

Lastly, in case of sufficient data and a hardware configuration, video-level representation and spatio-temporal feature aggregation steps can be trained all together. However, due to the high network complexity and lack of data, these steps are considered/learned separately in this study.

\section{Implementation Details}

All trainable parameters in the model are optimized with a gradient-based stochastic Adam solver~\cite{kingma2014adam}. Mini-batch size and learning rate are set to 64 and 0.001 respectively. The learning rate is decreased exponentially with the factor of 0.1 for every 40K iteration. Also, we set $\beta_1$ value in~\cite{kingma2014adam} to $0.7$ to reduce the oscillations in learning phase. Number of iterations varies from 50K to 200K depending on the loss changes between the training and validation sets for each emotion type.

Our implementation uses both Caffe~\cite{jia2014caffe} and Tensorflow~\cite{abadi2016tensorflow} frameworks. For CBP, we adapt the open source implementation\footnote{\url{https://github.com/ronghanghu/tensorflow_compact_bilinear_pooling}} into our code. Also, NetVlad is taken from the implementation\footnote{\url{https://github.com/antoine77340/LOUPE}}. Models are trained and evaluated on a single NVIDIA Tesla K40 GPU card. 

\vspace{0.2cm}
\noindent
\textbf{Parameter Configuration}: Parameter configurations of the proposed setup are given in accordance with empirical inference. $T$ and $K$ are set to 3 and 5, respectively. Therefore, the spatio-temporal method encapsulates 150 ms frame interval to extract a compact representation. Similar optimum parameter setup can be observed in~\cite{wang2011action} for action recognition problem. Also,~\cite{iwasaki2016hiding} shows that 150 ms frame interval is reasonable to assess micro-level emotional changes for real-fake expression prediction. 

Moreover, M is 36 (i.e. $6\times 6$), since each cropped face is fed into the precomputed CNN model as a $224\times 224$ spatial resolution image.  

\section{Experiments}

\vspace{0.2cm}
\noindent
\textbf{Real-Fake Expression Dataset}: The dataset consists of 600 videos~\cite{ofodile2017automatic, challenge}. Their lengths vary between 3-4 seconds and 50 different subjects perform 6 universal facial emotion types (anger, happiness, surprise, disgust, contentment, sadness) with genuine (real) and deceptive (fake) expressions. Moreover, the dataset is collected at high fps, i.e. 100fps. Fig.~\ref{fig:data} illustrates both genuine and deceptive facial expressions of people for each emotion type.

\begin{table}[t]
\begin{center}

\begin{tabular}{ l c }
\hline \hline
Model & MAP \\ 
\hline 
CBP & 53.33 \% \\ 
PCA+CBP & 46.66 \% \\  
\hline
NetVLAD & 56.66 \% \\ 
PCA+NetVLAD &  55.00 \% \\  
\hline 
RNN+CBP & 61.66 \% \\ 
PCA+RNN+CBP &  58.33 \% \\  
\hline
CBP+RNN+CBP (our) & \textbf{68.33} \%  \\
PCA+CBP+RNN+CBP (our) & 63.33 \%  \\
\hline
 CBP+RNN+NetVLAD (our) & 65.00 \%  \\
 PCA+CBP+RNN+NetVLAD (our) & 61.66 \%  \\
\hline \hline
\end{tabular}

\vspace{0.2cm}
\caption{Evaluation of individual methods in different sequential combinations.}
\label{tab:eval}
\end{center}
\end{table}

This dataset is divided into training, validation and test sets as 480, 60 and 60 videos respectively. Since we do not have access to the test labels, all results (except the official test accuracy at the end of this section) are reported on the validation set. Mean Average Presion (MAP) metric is used to evaluate the performances.

\vspace{0.2cm}
\noindent
\textbf{Model Evaluation}:  In the experiments, we first evaluate the performances of individual methods in different sequential combinations. For instance, in Table~\ref{tab:eval}, CBP+RNN+CBP indicates that after the extraction of deep features from faces, CBP is initially utilized to aggregate the features from multiple-frames. Later, RNN is used to capture short-temporal structure and lastly, CBP encodes the final video-level representation. 

To make a fair comparison, the dimension of the final video-level representation is fixed to 512 for all methods. Furthermore, cluster size for NetVlad is set to 32 to balance the complexity and accuracy.

From the results, we can observe that considering spatio-temporal interdependencies of deep features improves the accuracy compared to the methods merely exploiting feature pooling. Moreover, the use of PCA-like dimension reduction on deep features affects the accuracy adversely. Main reason can be explained as since dataset is very limited, estimated models can overfit to the training data. Similarly, using NetVlad in the video-level representation with RNN yields lower accuracy compared to CBP due to a small number of training data. 

Although exploiting the coarse frame-level features with RNN (i.e. RNN+CBP) still improves the accuracy over the feature pooling techniques (i.e. CBP or NetVlad), it obtains inadequate results compared to the spatio-temporal methods that use relaxed multi-frame feature aggregation before the temporal structure estimation  (i.e. CBP+RNN+CBP).

\begin{table}[t]
\begin{center}

\begin{tabular}{ l c c}
\hline \hline
Emotion & NetVLAD & CBP+RNN+CBP (our) \\ 
\hline 
Anger            & 50.00 \% & \textbf{80.00} \%  \\ 
Happiness     & 60.00 \% & \textbf{70.00} \%   \\ 
Surprise         & \textbf{70.00} \% & 60.00 \%  \\ 
Disgust          & 50.00 \% & \textbf{70.00} \%   \\ 
Contentment & \textbf{60.00} \% & \textbf{60.00} \%  \\ 
Sadness         & 50.00 \% & \textbf{70.00} \%  \\ 
\hline 
Average         & 56.66 \% & \textbf{68.33} \% \\
\hline \hline
\end{tabular}

\vspace{0.2cm}
\caption{Impact of spatio-temporal aggregation for individual emotion types. MAP scores are reported on the validation set. }
\label{tab:impact}
\end{center}
\end{table}

\begin{table}[b]
\begin{center}

\begin{tabular}{ l c c c }
\hline \hline
Emotion & Train & Val. & Test \\ 
\hline 
Anger            & 96.25 \% & 80.00 \% & - \\ 
Happiness     & 96.25 \% & 70.00 \% & - \\ 
Surprise         & 92.50 \% & 60.00 \% & - \\ 
Disgust          & 96.25 \% & 70.00 \% & - \\ 
Contentment & 92.50 \% & 60.00 \% & - \\ 
Sadness         & 93.75 \% & 70.00 \% & - \\ 
\hline 
Average         & 94.58 \% & 68.33 \% & 65.00  \% \\
\hline \hline
\end{tabular}

\vspace{0.2cm}
\caption{Official MAP scores for our method (i.e. CBP+RNN+CBP) on real-fake expression prediction dataset.}
\label{tab:result}
\end{center}
\end{table}

\vspace{0.2cm}
\noindent
\textbf{Impact of Spatio-Temporal Aggregation}:  In this part, we discuss individual emotional type accuracies for NetVLAD and our relaxed spatio-temporal method (i.e. the combination of CBP+RNN+CBP). In Table~\ref{tab:impact}, the performances of each individual emotion are reported for these two methods. From these results, for some emotional types such as 'surprise' and 'contentment', capturing only feature interdependencies yields the same/better performances compared to the spatio-temporal model.  Interestingly, when we investigate these training videos visually, we can observe that spatial emotional variations on face parts can be sufficiently distinct to categorize a person expression as real-fake, rather than exploiting the temporal structure/consistency (Note that we have no experience on this field).

On the other hand, the spatio-temporal method outperforms the feature aggregation technique (NetVlad) especially on 'anger', 'happiness', 'disgust' and 'sadness' due to the fact that exaggerated emotional changes on face parts are one of the useful patterns for the problem as we mentioned early. Note that our method can learn these patterns from data by itself.

\vspace{0.2cm}
\noindent
\textbf{Official Test Performance}: In Table~\ref{tab:result}, the official training, validation and test MAP performances for our method on real-fake expression prediction challenge are given. Since training data is small for the problem, overall training performance is quite high compared to the validation and test sets. That's why, SVM is a suitable option in the classification phase due to its generalization capacity from a small number of data.

\section{Conclusion}

In this study, we introduce a relaxed and learnable spatio-temporal feature aggregation method. Basically, it is able to capture the temporal structure between deep features and their spatial interdependencies. We explained the advantages on the feature aggregation step in detail. To demonstrate the effectiveness of the method, we tackled the real-fake emotion prediction problem. Our method achieves 65$\%$ MAP score on the test dataset and there is only one misclassified decision with the best result in the Chalearn Challenge. In future, we plan to apply our method on action/event recognition which has relatively larger training data to show its performance on different problems.

\section{Acknowledgment}
The authors are grateful to NVIDIA Corporation for the donation of NVIDIA Tesla K40 GPU card used for this research.

{\small
\bibliographystyle{ieee}
\bibliography{egbib}
}

\end{document}